\documentclass[letterpaper, 10 pt, conference]{ieeeconf}  % Comment this line out if you need a4paper

\IEEEoverridecommandlockouts                              % This command is only needed if 
\usepackage{algorithm}
\usepackage{algorithmic}
\usepackage{subcaption}
\usepackage{amsmath} 
\usepackage{amsfonts}       % blackboard math symbols
\usepackage{amssymb}
\usepackage{arydshln} 
\usepackage[percent]{overpic}
\usepackage[bookmarks=true,colorlinks=true]{hyperref}
\usepackage{hyperref}
\usepackage{url}
\usepackage{multirow}
\usepackage{graphicx}
\usepackage{varioref}
\usepackage{tabularx}
\usepackage{booktabs}
\usepackage{dsfont}
\usepackage{xcolor}         % colors
\usepackage{makecell}

\newcommand{\ie}{\textit{i.e.}}
\newcommand{\eg}{\textit{e.g.}}
\newcommand{\ours}{PLARE}
\newcommand{\fullours}{\textbf{P}referece-based \textbf{L}earning from Vision-L\textbf{A}nguage Model without \textbf{R}eward \textbf{E}stimation}

\newcommand{\stdv}[1]{\scalebox{0.8}{$\pm$~#1}}

\title{\LARGE \bf
Policy Learning from Large Vision-Language Model Feedback\\ Without Reward Modeling
}

\author{Tung M. Luu, Donghoon Lee, Younghwan Lee, and Chang D. Yoo$^*$ % <-this % stops a space
\thanks{$^*$Corresponding author: Chang D. Yoo} %
\thanks{All authors are with the School of Electrical Engineering, KAIST (Korea Advanced Institute of Science and Technology), Daejeon, Republic of Korea. \{tungluu2203,dh\_lee99,youngh2,cd\_yoo\}@kaist.ac.kr}%
\thanks{This work was partially supported by Institute for Information \& communications Technology Planning \& Evaluation (IITP) grant funded by the Korea government(MSIT) (No.RS-2021-II211381, Development of Causal AI through Video Understanding and Reinforcement Learning, and Its Applications to Real Environments) and Institute of Information \& communications Technology Planning \& Evaluation (IITP) grant funded by the Korea government(MSIT) (No. RS-2022-II0951, Development of Uncertainty-Aware Agents Learning by Asking Questions).}
}

\begin{document}

\maketitle
\thispagestyle{empty}
\pagestyle{empty}

%%%%%%%%%%%%%%%%%%%%%%%%%%%%%%%%%%%%%%%%%%%%%%%%%%%%%%%%%%%%%%%%%%%%%%%%%%%%%%%%
\begin{abstract}

Offline reinforcement learning (RL) provides a powerful framework for training robotic agents using pre-collected, suboptimal datasets, eliminating the need for costly, time-consuming, and potentially hazardous online interactions. This is particularly useful in safety-critical real-world applications, where online data collection is expensive and impractical. However, existing offline RL algorithms typically require reward labeled data, which introduces an additional bottleneck: reward function design is itself costly, labor-intensive, and requires significant domain expertise. In this paper, we introduce \ours{}, a novel approach that leverages large vision-language models (VLMs) to provide guidance signals for agent training. Instead of relying on manually designed reward functions, \ours{} queries a VLM for preference labels on pairs of visual trajectory segments based on a language task description. The policy is then trained directly from these preference labels using a supervised contrastive preference learning objective, bypassing the need to learn explicit reward models. Through extensive experiments on robotic manipulation tasks from the MetaWorld, \ours{} achieves performance on par with or surpassing existing state-of-the-art VLM-based reward generation methods. Furthermore, we demonstrate the effectiveness of \ours{} in real-world manipulation tasks with a physical robot, further validating its practical applicability.

\end{abstract}

%%%%%%%%%%%%%%%%%%%%%%%%%%%%%%%%%%%%%%%%%%%%%%%%%%%%%%%%%%%%%%%%%%%%%%%%%%%%%%%%
\section{INTRODUCTION}
Reinforcement learning (RL) has demonstrated remarkable success across various decision-making domains, including games \cite{vinyals2019grandmaster,perolat2022mastering}, autonomous systems \cite{bellemare2020autonomous,kaufmann2023champion}, and robotics \cite{kalashnikov2018scalable,luu2021hindsight,chen2022towards}. A key factor behind these successes is the design of well-crafted reward functions that map transitions to meaningful scalar rewards. However, crafting effective reward functions often requires substantial human effort and domain expertise. Moreover, RL agents are prone to reward hacking, where they exploit reward structures to achieve high returns while exhibiting unintended or undesirable behaviors, which may lead to dangerous consequences \cite{hadfield2017inverse,skalse2022defining}. 

In this work, we explore the potential of large Vison-Language Models (VLMs) in addressing these challenges. VLMs have shown remarkable ability to capture complex relationships between visual and textual information. When combined with Large Language Models (LLMs), they have been effectively utilized across a wide range of multimodal tasks, from visual question answering \cite{zhu2023minigpt,yoon2023hear,bao2024autobench} to robotics \cite{kim2024openvla,venuto2024code}. LLMs have also been leveraged to enhance skill exploration for RL \cite{du2023guiding,zhang2023bootstrap} and to generate executable code that represent policies \cite{liang2023code} or planners \cite{singh2023progprompt}. Given their ability to accurately describe tasks and provide high-level judgments from visual inputs, large VLMs present a promising opportunity for improving robotic agents that rely on image-based or multimodal observations. Concretely, in this work, we leverage large VLMs to generate guidance signals based on language task descriptions and the agent's visual observations. These signals are then used to train policies to perform manipulation tasks, reducing the extensive human effort required to manually specify reward functions.

\begin{figure}[t]
    \centering
    \includegraphics[width=\columnwidth]{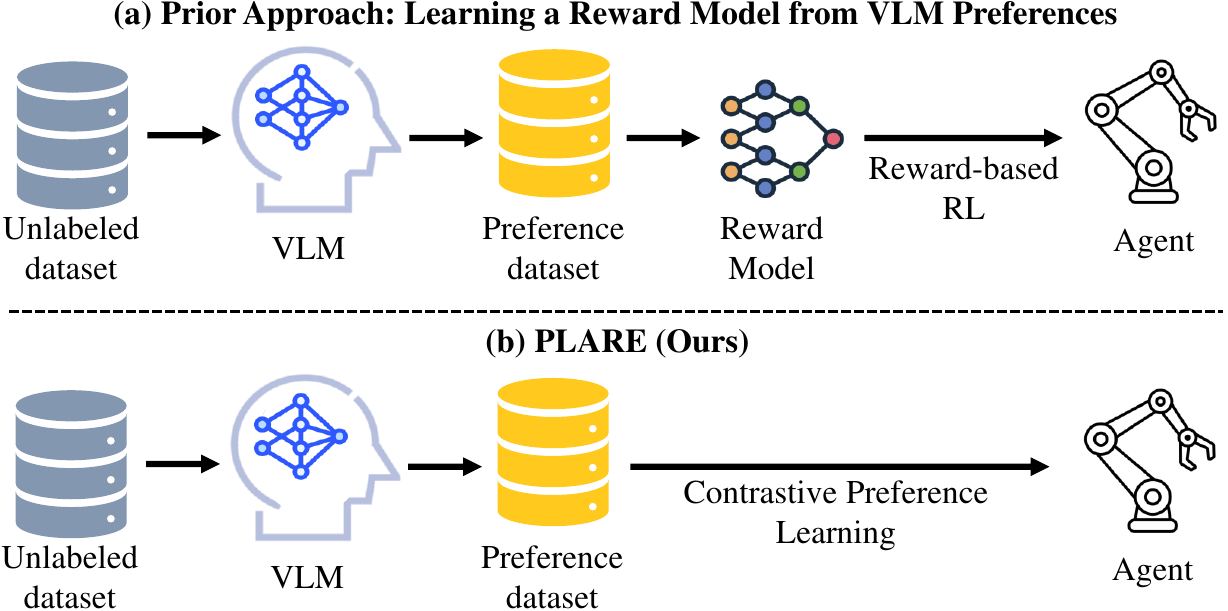}
    \caption{\ours{} utilizes a VLM to generate preference labels for pairs of trajectory segments sampled from an unlabeled dataset. These preferences are then directly leveraged for policy optimization via a contrastive learning objective, eliminating the need for an intermediate reward modeling step.}
    \label{fig:intro}
    % \vskip -0.25in
\end{figure}

A direct approach to achieving this goal is to use pretrained VLMs, such as CLIP \cite{radford2021learning}, to compute similarity scores between visual observations and task descriptions within a learned embedding space, as explored in \cite{sontakke2024roboclip,rocamonde2024vision}. However, these similarity-based reward signals are often too coarse for fine-grained tasks like robotic manipulation, as their reliability heavily depends on the quality of task descriptions and the alignment of agent observations with the pretraining data distribution \cite{sontakke2024roboclip,rocamonde2024vision,wang2024rl}. 
An alternative line of research leverages LLMs/VLMs to generate structured reward functions in the form of executable code \cite{yu2023language,xie2024text2reward,venuto2024code}. While effective in simulation, these approaches require access to the environment’s underlying code, making them impractical for real-world deployment. Additionally, evaluating such reward functions typically involves multiple full RL training cycles, leading to high computational costs.
More recently, a different paradigm has emerged, where LLMs/VLMs serve as teachers by providing preference feedback. For instance, \cite{wang2024rl,venkataraman2024real} build upon the standard preference-based RL framework \cite{christiano2017deep,lee2021b}, using VLMs to generate the preference labels over two visual observations. These preferences are then used to train a reward model, which is subsequently used for policy optimization via an RL algorithm.
While this approach has shown competitive performance across various manipulation tasks, it introduces several challenges. First, relying on a learned reward model introduces cascading prediction errors that propagate from the reward function to the critic and ultimately to the actor, leading to high variance in policy performance \cite{hejna2023inverse}. Second, VLM-generated feedback is susceptible to hallucinations \cite{yoon2022information,zhang2023siren,guan2024task}, further compounding error propagation and reducing the reliability of learned reward models. Finally, RL algorithms are already inherently delicate, often involving multiple interacting components that require careful hyperparameter tuning. Adding a separate reward model further increases computational overhead, especially for large models \cite{early2022non,kim2023preference}, and intensifies the challenge of tuning. This added complexity makes this approach less scalable for real-world deployment. 

To this end, we introduce \ours{} (\fullours{}), a novel framework that harnesses large VLMs to generate preference feedback for training robotic agents on new tasks. Specifically, given an unlabeled dataset, our method queries large VLMs (\ie, Gemini \cite{reid2024gemini}) to generate preferences between two visual trajectory segments sampled from this dataset, based solely on the human-written language task description. We then build upon the recently introduced contrastive preference learning framework \cite{hejna2024contrastive} to directly optimize policies from the preference dataset, bypassing the need for explicitly learning a reward model (Figure \ref{fig:intro}). This enables rapid adaptation to new tasks while requiring only a natural language task description, significantly reducing the human effort involved in reward function design. We evaluate \ours{} on a diverse set of manipulation tasks in both simulated and real-world environments, demonstrating competitive performance across tasks. Our main contributions are summarized as follows:
\begin{enumerate}
    \item We introduce \fullours{} (\ours), a novel framework that enables agents to learn tasks using only language task descriptions and the agent's visual observations, this reduces the effort for manually designed reward functions.
    \item We demonstrate the effectiveness of \ours{} on a set of manipulation tasks from MetaWorld \cite{yu2020meta}, achieving performance that surpasses prior VLM-based reward generation methods. Additionally, we provide analyses to better understand the impact of our approach. 
    \item We validate \ours{} on real-world manipulation tasks, demonstrating its ability to learn effective policies from suboptimal datasets and outperform baselines. The code is available at: \url{https://github.com/tunglm2203/plare}.
\end{enumerate}

\section{RELATED WORK}

\subsection{VLMs as Reward Functions}
A common approach to leveraging vision-language models (VLMs) for reward generation in reinforcement learning (RL) is to assess the alignment between visual observations and textual task descriptions. For instance, MineDojo \cite{fan2022minedojo} fine-tunes a CLIP model on a large corpus of Minecraft gameplay videos, producing MineCLIP, which is then used to quantify how well an agent’s observation aligns with a given task description (\ie, ``collect wood'') and uses this score as a reward signal. While such VLM-based reward functions can effectively guide agents in complex, visually rich environments, fine-tuning on domain-specific data (such as Minecraft or robotics) is often computationally expensive, particularly for large models. To address this, recent work has explored using pretrained VLMs to generate rewards in a zero-shot manner \cite{sontakke2024roboclip,rocamonde2024vision}. However, these similarity-based reward signals are often noisy and inconsistent, and their accuracy heavily depends on the task description and the alignment between visual observations and the data on which the VLM was pretrained \cite{ma2023liv,wang2024rl}. Another promising approach is to directly prompt large VLMs to generate preference or rating labels \cite{wang2024rl,venkataraman2024real,luu2025enhancing}, which are then used to train a reward model for RL optimization. In contrast, our method directly train the policies from VLM preference feedback without requiring reward modeling. This reduce the burden in parameter tuning for training reward models as well as overhead for large reward models.

\begin{figure*}[ht]
    \centering
    \includegraphics[width=\textwidth]{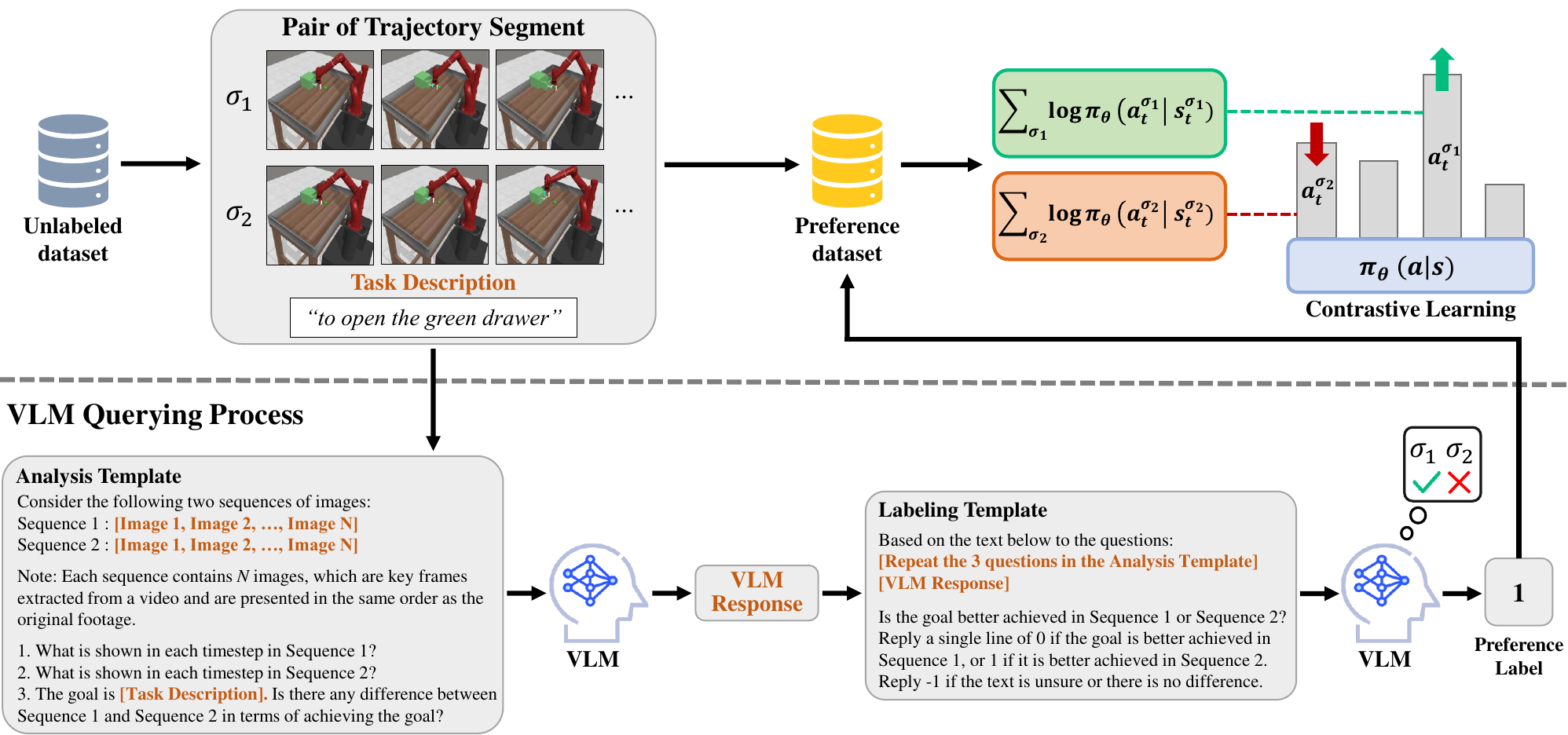}
    \caption{\textbf{Overview of \ours{}}: Given an unlabeled, reward-free dataset, \ours{} samples two trajectory segments and queries a VLM teacher for a preference label based on the language task description. Notably, only visual observations from each segment are used for querying the VLM. The selected segments, along with their assigned preference labels, are then stored in the preference dataset. Finally, the policy is learned from this dataset using a contrastive learning objective.}
    \label{fig:overview}
    % \vskip -0.2in
\end{figure*}

\subsection{RL In The Absence Of Reward Functions}
There is a rich body of research on policy learning in the absence of reward functions. One widely studied approach is imitation learning (IL), which seeks to learn policies from expert demonstrations. Behavior cloning \cite{pomerleau1988alvinn}, a common IL method, formulates policy learning as supervised learning over state-action pairs and typically relies on large datasets of expert trajectories \cite{brohan2022rt, lynch2023interactive,luu2024predictive}. However, obtaining such large-scale demonstrations is often expensive to collect. An alternative IL strategy is inverse RL (IRL) \cite{ng2000algorithms,ho2016generative}, which aims to recover a reward function from expert demonstrations. Various IRL methods have been developed to infer more effective reward functions, including nonlinear IRL \cite{levine2011nonlinear,finn2016guided} and adversarial IRL \cite{ho2016generative,fu2017learning}. Despite its potential, IRL methods heavily depend on high-quality demonstrations, which can be difficult and expensive to obtain for complex tasks \cite{zhang2021confidence,brown2019extrapolating}. Recent advances have explored using powerful diffusion models \cite{ho2020denoising,pham2025mdsgen,ton2025taro} to model expert behavior and generate reward signals \cite{huang2024diffusion}, offering an alternative to manually designed rewards. Other works have instead leveraged human comparative feedback to learn reward functions, a paradigm known as preference-based RL (PbRL) \cite{christiano2017deep, lee2021b}. Such comparative feedback often requires less human effort than expert demonstrations, as it only asks humans to provide relative comparisons between trajectory pairs rather than directly performing the task. Recent works further introduce simplified PbRL approaches, which directly train policies from preference feedback in a supervised manner \cite{hejna2023inverse,an2023direct,hejna2024contrastive}. However, a major limitation of PbRL is its dependence on extensive human participation to generate the necessary feedback. Our method addresses this limitation by leveraging VLMs to generate preference feedback, eliminating the need for human-provided comparisons while retaining the benefits of preference-based learning.

\section{BACKGROUND}

\subsection{Reinforcement Learning}
A decision-making problem is typically modeled by a Markov decision process (MDP), which can be described as a tuple of $\mathcal{M}=(\mathcal{S}, \mathcal{A}, p, R, \gamma, H)$, where $\mathcal{S}$ is a state space, $\mathcal{A}$ is an action space, $p(s'|s, a)$ is the state transition dynamics, $R(s, a)$ is the reward function, $\gamma\in[0, 1]$ is the discount factor, and $H$ is the time horizon. The goal of reinforcement learning (RL) algorithm is to learn a policy $\pi$ that maximize the expected return $\mathbb{E}_{\pi}[\sum_{t=0}^H\gamma^{t}R(s_t, a_t)]$. In this work, we assume that the true reward $R$ is unknown. However, we have access to a unlabeled, reward-free dataset, $D = \{(s_0^j, a_0^j, s_1^j, a_1^j, \cdots)\}_{j=1}^{M}$, which contains $M$ trajectories collected by an unknown policy, such as human experts or scripted policies, attempting to complete the task.

\subsection{Offline Preference-based Reinforcement Learning}
In this work, we consider the offline preference-based RL (PbRL) framework \cite{hejna2023inverse,kim2023preference}, where the agent is allowed to query a teacher for preference feedback to obtain preference labels over pairs of trajectory segments. Specifically, we first construct a comparison dataset by randomly sampling segment pairs from an unlabeled dataset $D$, forming the comparison set $D_{c}=\{(\sigma_1^i, \sigma_2^i) | \sigma=(s_0, a_0, \cdots, s_L, a_L), \sigma\in D\}_{i=1}^N$, which contains $N$ pairs of $L$-length segments. Next, the teacher assigns a ternary label $y$ to each segment pair $(\sigma_1, \sigma_2) \in D_c$: $y=0$ indicates that $\sigma_1$ is preferred over $\sigma_2$, $y=1$ indicates the opposite preference, and $y=0.5$ signifies equal preference. This process results in the preference dataset $D_{pref}=\{(\sigma_1^i, \sigma_2^i, y_i)\}_{i=1}^N$. Conventionally, preference labels are used to learn the unknown reward function, which is then optimized via RL. Recent works propose performing offline PbRL without an explicit reward model by directly optimizing policies from preference labels \cite{hejna2023inverse,an2023direct,hejna2024contrastive}. In our framework, we adopt contrastive preference learning (CPL) \cite{hejna2024contrastive} as the preference learning algorithm. Mathematically, given the preference dataset $D_{pref}$, the policy $\pi_{\theta}$ is optimized as follows:

% {\small
\begin{equation}\label{eq:cpl}
    \mathcal{L}(\theta) = -\mathbb{E}_{D_{pref}}\left[(1-y)h(\pi_{\theta}, \sigma_1, \sigma_2) + yh(\pi_{\theta}, \sigma_2, \sigma_1)\right]
\end{equation}
% }
% \noindent
where $h(\pi_{\theta}, \sigma_1, \sigma_2)$ is computed as:

% {\small
\begin{equation}
    \log\frac{e^{\sum_{(s_t, a_t)\in\sigma_1}\gamma^t\alpha\log{\pi_{\theta}(a_t|s_t)}}}{e^{\sum_{(s_t, a_t)\in\sigma_1}\gamma^t\alpha\log{\pi_{\theta}(a_t|s_t)}} + e^{\lambda\sum_{(s_t, a_t)\in\sigma_2}\gamma^t\alpha\log{\pi_{\theta}(a_t|s_t)}}}
\end{equation}
% }
where, $\alpha$ and $\lambda$ are hyperparameters.

\section{METHOD}
An overview of \ours{} is presented in Figure \ref{fig:overview}. We first leverage a VLM (such as Gemini \cite{reid2024gemini}) to generate preference labels, constructing a preference dataset. The policy is then optimized using a supervised learning objective based on the generated dataset. The detailed procedure of \ours{} is provided in Algorithm \ref{alg:pseudo_code}.

\subsection{Preference Feedback from VLM}
For each sampled pair of trajectory segments, we query the VLM to determine which sequence of images better performs the task according to the language task description (Algorithm \ref{alg:pseudo_code}, lines 5-6). To obtain feedback, we perform querying in two stages. In the first stage, we construct a prompt containing two sequences of images along with questions that request descriptions of each sequence and a comparison between them. In the second stage, we use the VLM’s response from the first stage to obtain the final preference label for the two sequences. The detailed prompt is shown in Figure \ref{fig:overview} (bottom). Our querying process is inspired by \cite{wang2024rl}; however, instead of using a single image, we provide a sequence of images and a modified textual prompt to analyze each timestep. Finally, we store the compared segments along with their preference labels in the preference dataset $D_{pref}$ (Algorithm \ref{alg:pseudo_code}, line 7). 

We fix the number of images queried from the VLM to three per segment, selecting the first, middle, and last frame. While using fewer images per segment (\eg, single frame) reduces query time, it lacks the granularity needed to represent agent behavior accurately. On the other hand, using more images (\eg, 30) provides a more detailed view of the agent’s actions but significantly increases query time due to the larger input size. Through preliminary experiments, we find that selecting three representative frames offers an effective trade-off, minimizing the number of queries while preserving enough temporal context for the VLM to make accurate preference judgments. Although more advanced methods could be used to select informative frames within each segment \cite{zhang2020atari, yoon2023scanet}, we find that three uniformly spaced frames strike a practical and effective balance.

\begin{algorithm}[t]
% \small
\caption{\ours{}}
\label{alg:pseudo_code}
\begin{algorithmic}[1]
    \STATE \textbf{Input}: Unlabeled dataset $D$, segment length $L$, number of queries $N$, and text description of task goal $l$.
    \STATE \textbf{Initialize}: Policy $\pi_{\theta}$, preference buffer $D_{pref} \leftarrow \emptyset$.
    \STATE // \texttt{GENERATE PREFERENCE}
    \FOR{$i=1$ to $N$}
        \STATE Randomly sample $(\sigma_1, \sigma_2) \sim D$ with length of $L$
        \STATE Query VLM with $(\sigma_1, \sigma_2)$ and task goal $l$ for label $y$
        \STATE Update $D_{pref} \leftarrow D_{pref} \cup \{(\sigma_1, \sigma_2, y)\}$
    \ENDFOR
    
    \STATE // \texttt{POLICY LEARNING}
    \FOR{each iteration}
        \STATE Sample random batch $\{(\sigma_1, \sigma_2, y)_j\}_{j=1}^B \sim D_{pref}$
        \STATE Update the policy $\pi_{\theta}$ with Equation (\ref{eq:cpl})
    \ENDFOR
\end{algorithmic}
\end{algorithm}

\subsection{Policy Training}\label{subsec:policy_learning}

Given the collected preference dataset, we optimize the policy using the CPL objective as in Equation (\ref{eq:cpl}). Notably, this training paradigm is entirely supervised, relying solely on preference feedback without RL optimization. While the original CPL objective accounts for equal preference cases (\ie, $y=0.5$), we find that excluding these samples often leads to improved performance, as demonstrated in our experiments. Moreover, since VLM-generated preference labels may by noisy due to hallucination \cite{zhang2023siren,guan2024task}, we tune the dropout rate during policy training, leveraging it as an effective regularizer for learning from noisy labels \cite{jindal2017learning,chen2022compressing}.

\section{EXPERIMENTS}

\begin{figure}[ht]
    \centering
    \includegraphics[width=0.75\columnwidth]{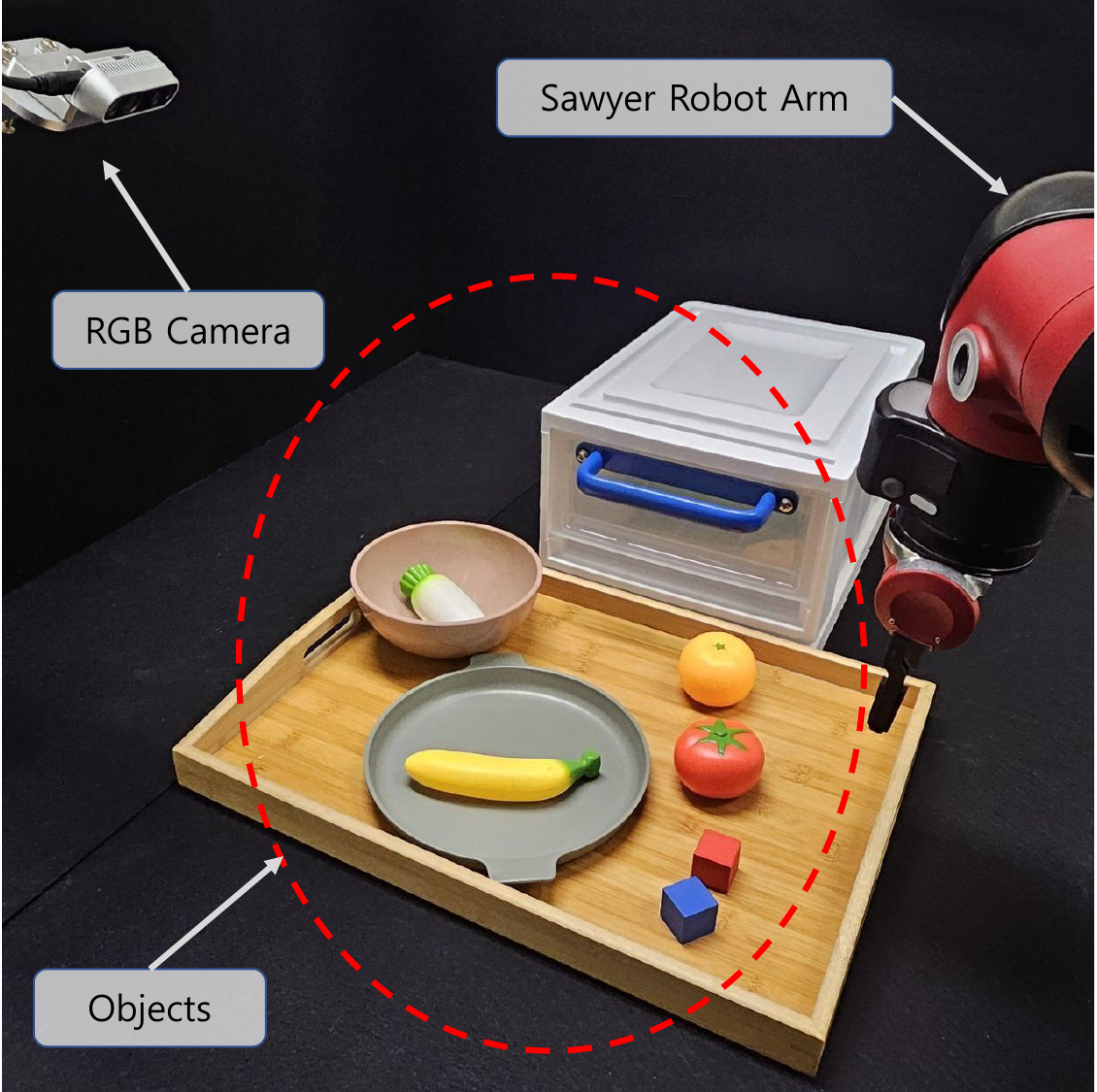}
    \caption{Workspace setup for our 7-Dof Sawyer Robot arm with RGB image-based control.}
    \label{fig:realrobot}
    % \vskip -0.1in
\end{figure}

\begin{table*}[ht]
\centering
\renewcommand{\arraystretch}{1.21} % For ArXiv set to 1.21
\resizebox{0.92\textwidth}{!}{
\begin{tabular}{@{}lccccc@{}}
&  \includegraphics[width=1.75cm]{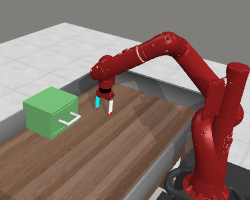}  &
   \includegraphics[width=1.75cm]{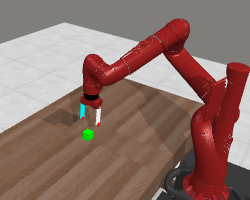}   &
   \includegraphics[width=1.75cm]{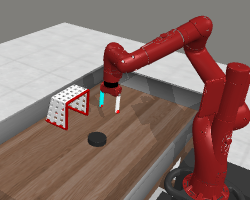}  & 
   \includegraphics[width=1.75cm]{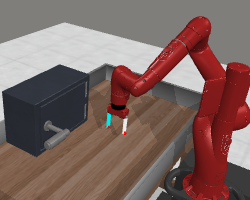}    & \\

& Drawer Open & Sweep Into & Plate Slide  &  Door Open   & Average \\ 
\midrule
 
IQL-CLIP \cite{rocamonde2024vision}
& 63.0 \stdv{9.2}    & 62.0 \stdv{12.8}  & 47.0 \stdv{7.7}   & 66.0 \stdv{10.8} & 59.50  \\

IQL-RoboCLIP \cite{sontakke2024roboclip}
& 69.0 \stdv{11.5}   & 50.0 \stdv{11.8}  & 33.0 \stdv{1.7}   & 48.0 \stdv{11.0} & 50.00  \\

RL-VLM-F \cite{wang2024rl}
& 67.4 \stdv{7.1}   & \textbf{65.1 \stdv{7.7}}  & 49.0 \stdv{4.7}   & 65.2 \stdv{10.3} & 61.68  \\

\ours{} (Ours)
& \textbf{84.9 \stdv{6.3}}  & \textbf{65.5 \stdv{7.1}}   & \textbf{58.7 \stdv{6.5}}  & \textbf{70.9 \stdv{6.7}} & \textbf{70.00}  \\

\midrule
BC  
& 61.5 \stdv{3.7}    & 49.3 \stdv{2.1}  & 39.1 \stdv{2.5}   & 57.5 \stdv{3}  & 51.85  \\

CPL (Oracle)
& 87.0 \stdv{5.2}   & 84.0 \stdv{2.8}  & 57.0 \stdv{4.5}   & 85.0 \stdv{9.9} & 78.25  \\
\bottomrule
\end{tabular}
}
\caption{Success rate (in percent) of all methods across four manipulation tasks from MetaWorld. Each method is evaluated using four random seeds. We report the maximum average performance across seeds over an 8-checkpoint, 200-episode evaluation window, following the protocol in \cite{hejna2024contrastive}. Bolded values indicate results within 1\% of the top-performing method (exclude BC and Oracle). For CPL, we rerun its provided implementation to ensure a fair comparison and find that it generally achieves better performance than reported in the original paper.}
\label{tab:sim_results}
% \vskip -0.1in
\end{table*}

\subsection{Experimental Setup}

We evaluate \ours{} on a suite of manipulation tasks from MetaWorld \cite{yu2020meta}, using unlabeled datasets from the offline PbRL benchmark \cite{hejna2024contrastive}. Additionally, we validate our approach in a real-world robotic manipulation setting using a 7-DOF Rethink Sawyer robot in the tabletop environment. The detailed tasks are as follows.

\noindent
\textbf{MetaWorld:} In this domain, the agent learns to generate low-level continuous control signals for a simulated Sawyer robotic arm, enabling it to interact with objects on a tabletop and achieve specified task objectives. Following \cite{hejna2024contrastive}, we evaluate methods on four manipulation tasks:

\begin{itemize}
    \item \textit{Drawer Open}: Pull open a drawer on the table.
    \item \textit{Sweep Into}: Sweep a green puck into a square hole on the table.
    \item \textit{Plate Slide}: Slide a plate into the goal on the table.
    \item \textit{Door Open}: Pull open a safe door on the table.
\end{itemize}

\noindent
\textbf{Real-World Robot Manipulation:} To assess the applicability of \ours{} in real-world robotic control, we implement vision-based manipulation tasks using a 7-DOF Rethink Sawyer robot  arm operating in a tabletop environment. The evaluation includes the following tasks:
\begin{itemize}
    \item \textit{Pickup Banana}: Grasp a banana from a plate and lift it.
    \item \textit{Drawer Open}: Pull open a white drawer on the table.
\end{itemize}

\begin{table*}[ht]
\small
\centering
\renewcommand{\arraystretch}{1.21} % Increase row height for readability
\resizebox{0.92\textwidth}{!}{
\begin{tabular}{l|c|c} % Three columns
    \toprule
    \textbf{Method} &\textbf{Task Name} & \textbf{Task Description} \\
    \midrule
    \multirow{6}{*}{CLIP} 
    & Drawer Open & the drawer is opened \\   
    & Sweep Into  & the green cube is in the hole \\
    & Plate Slide & the black plate is in the goal \\
    & Door Open   & the door of the safe is opened \\
    & Pickup Banana (Real) & the banana is grasped \\
    & Drawer Open (Real) & the door of the safe is opened \\
    \midrule
    
    \multirow{6}{*}{RoboCLIP} 
    & Drawer Open & robot opening green drawer \\   
    & Sweep Into & robot sweeping the green cube into the hole on the table \\
    & Plate Slide & robot pushing the black plate into the goal \\
    & Door Open & robot opening the safe door \\
    & Pickup Banana (Real) & robot picking up the banana \\
    & Drawer Open (Real) & robot opening the white drawer \\
    \midrule
    
    \multirow{6}{*}{RL-VLM-F / \ours{}} 
    & Drawer Open & to open the drawer \\   
    & Sweep Into & to minimize the distance between the green cube and the hole \\
    & Plate Slide & to move the black plate into the goal \\
    & Door Open & to open the safe door as large as possible \\
    & Pickup Banana (Real) & to grasp the banana from the plate and lift it upward \\
    & Drawer Open (Real) & to open the white drawer in the table \\
    \bottomrule
\end{tabular}
}
\caption{Language task descriptions for the tasks used in methods.}
\label{tab:mw_task_descriptions_clip}
\vskip -0.1in
\end{table*}

\noindent
The visualization of MetaWorld tasks is shown in Table \ref{tab:sim_results}, and the workspace setup for real-world tasks is presented in Figure \ref{fig:realrobot}. For MetaWorld tasks, we use the dataset\footnote{This dataset contains both low-dimensional states and their corresponding rendered images from the simulator.} provided by \cite{hejna2024contrastive}, which contains 2500 episodes with an approximate 50\% success rate. For real-world tasks, we collect 100 episodes via keyboard teleoperation, achieving an approximate 60\% success rate. In MetaWorld domain, the policy is learned using low-dimensional states for all methods. The low-dimensional state is a 35-dimensional vector obtained from the simulator, which includes the end-effector (EE) pose, gripper status, object pose, and goal position. In real-world domain, we use both low-dimensional states and visual observations for policy learning. The visual observation is an RGB image captured by an Intel RealSense D435i camera at a resolution of $480\times 480$, and resized to $224\times 224$. The low-dimensional state is a 9-dimensional vector, where the first six dimensions represent the Cartesian position and velocity of EE, the next dimension represents its yaw orientation, and the last two dimensions represent the gripper status in one-hot encoding (open/close). For visual feature extraction, we use R3M \cite{nair2022r3m} as the image encoder, producing a 512-dimensional image feature. This is concatenated with the low-dimensional state, resulting in a 521-dimensional vector as input to the policy. In both domains, the action represents the displacement of EE. Also, in the real robot setup, we constrain the EE to operate only with yaw orientation, and the robot is controlled continuously at 10 Hz. For all experiments, we use Gemini \cite{reid2024gemini} as the VLM teacher to obtain feedback. For MetaWorld tasks, we use $200\times200$ RGB images to query VLMs with $10k$ comparisons (\ie, $N=10k$ in Algorithm \ref{alg:pseudo_code}), while for real-world task, we use $224\times 224$ with $1k$ comparisons. The total querying time for $10k$ comparisons is approximately 10 hours. For evaluation, we use the task-defined success rate for MetaWorld tasks, while real-world robot tasks are evaluated by human assessment.

\subsection{Comparison with Previous Approaches}
We evaluate \ours{} against state-of-the-art (SoTA) VLM-based reward generation methods, which also leverage the agent's visual observations and language task descriptions for policy learning. Specifically, we consider the following baselines: (1) VLM-RMs \cite{rocamonde2024vision} (IQL-CLIP), which utilizes the CLIP model \cite{radford2021learning} to measure similarity between a rendered image and the task description; (2) RobotCLIP \cite{sontakke2024roboclip} (IQL-RoboCLIP), which employs the S3D model to measure similarity between video frames and the task description; and (3) RL-VLM-F \cite{wang2024rl}, a SoTA method that trains a reward model using VLM-generated feedback, which is subsequently used for RL optimization. To ensure a fair comparison, we use the same preference dataset for both RL-VLM-F and \ours{}. In RL-VLM-F, the reward model is trained using low-dimensional state inputs, whereas in IQL-CLIP and IQL-RoboCLIP, reward values are computed based on the task description and visual observations. The task descriptions for tasks are shown in Table \ref{tab:mw_task_descriptions_clip}. Additionally, we compare against behavior cloning (BC), a Learning from Demonstration method, and contrastive preference learning (CPL) \cite{hejna2024contrastive}, which is trained using ground-truth preference labels and serves as an oracle baseline. For all RL-based methods, we employ IQL \cite{kostrikov2022offline} as the underlying RL algorithm\footnote{We implement all baselines, including \ours{}, based on the CPL framework: \url{https://github.com/jhejna/cpl}. Default CPL hyperparameters are used for our method.}.

Results in Table \ref{tab:sim_results} demonstrate that \ours{} outperforms prior methods in three out of four tasks and is competitive with RL-VLM-F on \textit{Sweep Into}. We observe that RL-VLM-F surpasses IQL-CLIP and IQL-RoboCLIP in two out of four tasks. While RL-VLM-F was originally designed for online learning, its lower performance in the offline setting suggests that additional tuning is required for training its reward model, which is not needed in our method. Across all tasks, \ours{} achieves an average performance of 70.0, yielding a relative improvement of 13.5\% over SoTA. Notably, our method slightly outperforms the oracle baseline in the \textit{Plate Slide} task, highlighting the potential of VLMs in providing effective preference feedback.

\subsection{Ablation Studies}

In this section, we first analyze the impact of equal preference labels on policy learning. Next, as discussed in Section \ref{subsec:policy_learning}, we investigate the role of dropout as an effective regularizer for mitigating the effects of noisy labels. 

For the first analysis, we fix the dropout rate at 0.25 and compare policy training with and without equal preference labels. Table \ref{tab:ablation} (top) presents the proportion of equal preferences generated by the VLM in the preference dataset and the corresponding performance. The results consistently demonstrate that excluding equal preference labels improves performance, suggesting that ambiguous preference labels introduce uncertainty that hinders policy learning. This highlights the importance of filtering uncertain supervision, allowing the policy to focus on more decisive training signals.

\begin{table}[t]
% \vskip -0.05in
% \small
\centering
\renewcommand{\arraystretch}{1.21} % For ArXiv set to 1.21
\begin{tabular}{@{}lcccc@{}}
\toprule
Equal Pref. & Drawer Open & Sweep Into & Plate Slide  &  Door Open    \\ 
\midrule
\%Equal Pref. & 3.03\%    & 11.41\%  & 35.53\%   & 10.32\% \\
\cdashline{1-5}
Yes
& 80.0 \stdv{2.8}    & 62.0 \stdv{4.5}  & 51.1 \stdv{3.3}   & 69.5 \stdv{7.7} \\

No
& \textbf{82.1 \stdv{4.5}}   & \textbf{65.5 \stdv{7.1}}  & \textbf{56.3 \stdv{4.9}}   & \textbf{70.9 \stdv{6.7}}  \\
\midrule
\midrule

$p_{drop}$ & Drawer Open & Sweep Into & Plate Slide  &  Door Open    \\
\midrule
Accuracy & 70.02    & 69.42  & 80.67   & 70.01 \\
\cdashline{1-5}
 
0.1
& 80.0 \stdv{2.8}    & 63.0 \stdv{7.1}  & 54.5 \stdv{8.2}   & 67.2 \stdv{11.4} \\

0.25
& 82.1 \stdv{4.5}   & \textbf{65.5 \stdv{7.1}}  & 56.3 \stdv{4.9}   & \textbf{70.9 \stdv{6.7}}  \\

0.4
& \textbf{84.9 \stdv{6.3}}   & 63.1 \stdv{5.9}  & \textbf{58.7 \stdv{4.5}}   & 66.2 \stdv{10.0} \\

0.5
& 79.5 \stdv{6.1}   & 64.0 \stdv{8.5}  & 54.0 \stdv{6}   & 62.2 \stdv{4.5} \\
\bottomrule

\end{tabular}

\caption{Ablation study on the effect of using equal preference and dropout probability during policy learning. Bolded values indicate the best performance within each ablation. The best-performing configuration is used for the main results. Each experiment is conducted with four seeds.}
\label{tab:ablation}
\vskip -0.1in
\end{table}

In the second analysis, we exclude equal preferences and vary the dropout probability within $p_{drop} = \{0.1, 0.25, 0.4, 0.5\}$. Additionally, we use ground-truth preference labels to evaluate the accuracy of the preferences generated by the VLM. As shown in Table \ref{tab:ablation} (bottom), while the VLM can produce reasonable preferences, it still yields a moderate proportion of noisy labels (\eg, 20-30\% incorrect). This highlights the need for a mechanism to effectively learn from such noisy supervision \cite{cheng2024rime}. In this work, we adopt dropout as a simple yet effective regularization strategy. We observe that the optimal dropout probability differs across tasks, likely due to varying levels of label noise and its impact on different tasks. While some tasks benefit from higher dropout rates, others show diminishing returns, emphasizing the task-dependent nature of regularization in noisy preference learning. Overall, our results suggest that dropout effectively mitigates noise in preference labels, making it a suitable regularization technique for learning from imperfect preference labels.

\subsection{Computational Cost}

\begin{table}[ht]
\renewcommand{\arraystretch}{1.21} % For ArXiv set to 1.21
\centering
% \vskip -0.1in
\caption{Computational cost for $500k$ training steps on the Drawer Open task (simulation).}
\begin{tabular}{@{}lcc@{}}
    \toprule
    Method      & Parameters (M) & Runtime (hrs) \\ 
    \midrule
    RL-VLM-F    & 1.42   & 1.3   \\
    PLARE       & 0.28   & 0.4   \\
    % RL-VLM-F    & 1415688   & 1.3 hrs  \\
    % PLARE       & 283140   & 0.4 hrs  \\
    \bottomrule
\end{tabular}
\label{tab:computation_cost}
% \vskip -0.1in
\end{table}

We compare the computational cost of RL-VLM-F and \ours{} on a single machine equipped with an RTX 4090 GPU, as shown in Table \ref{tab:computation_cost}. The results indicate that \ours{} uses significantly fewer parameters and, as a result, achieves faster training times.

\subsection{Real Robot Evaluation}

\begin{figure}[ht]
    \centering
    \includegraphics[width=\columnwidth]{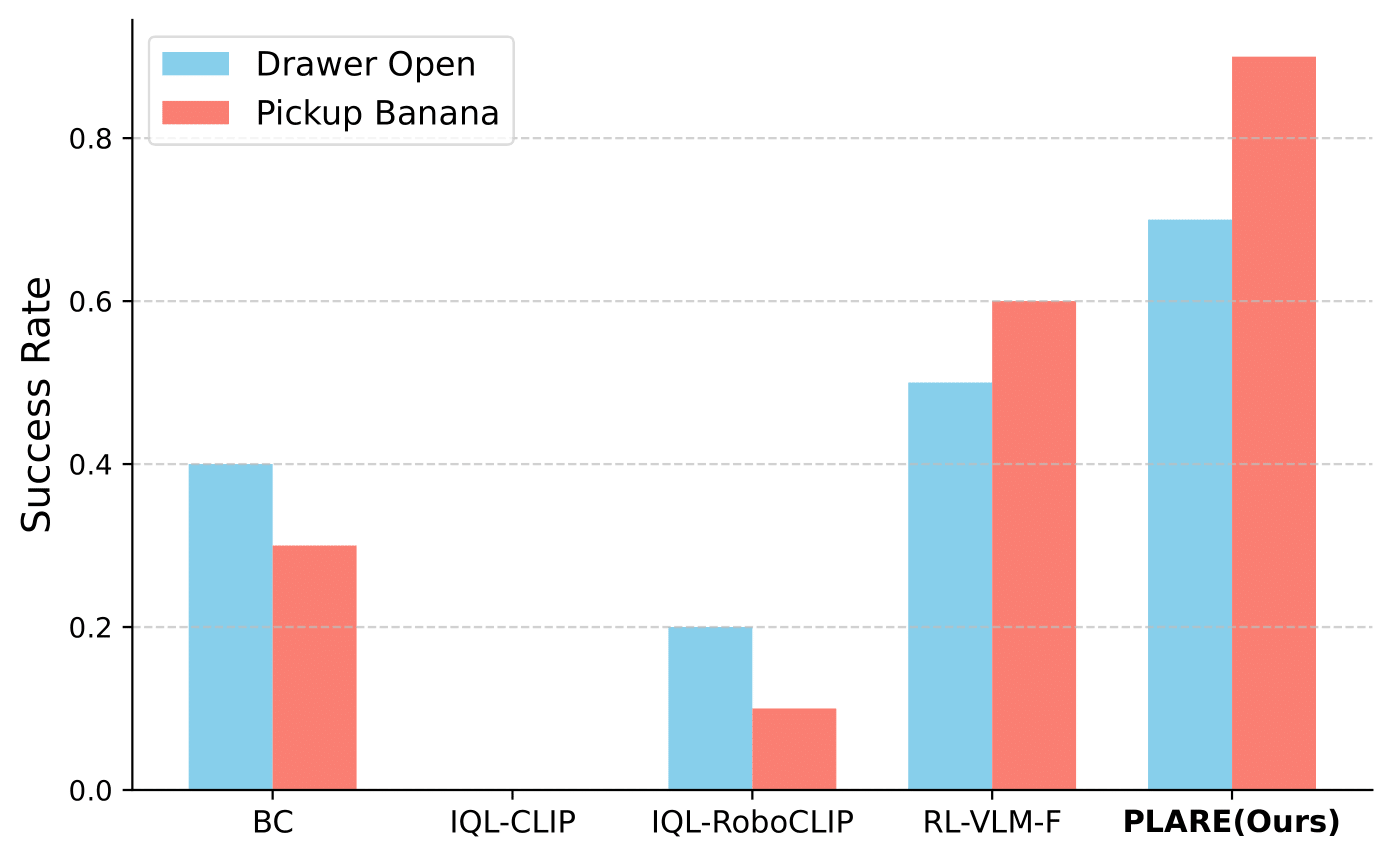}
    \caption{Success rates on the 2 real-world robot manipulation tasks in table-top environment.}
    \label{fig:realrobot_result}
    % \vskip -0.1in
\end{figure}

In this section, we demonstrate the effectiveness of \ours{} in two simple real-world robotic manipulation tasks. We use IQL as the RL algorithm for RL-based baselines: IQL-CLIP, IQL-RoboCLIP, and RL-VLM-F. Additionally, RL-VLM-F and \ours{} are used same preference dataset generated by the VLM. The hyperparameters for methods are same as MetaWorld experiments. Each method is evaluated over 10 trials, and the results are presented in Figure \ref{fig:realrobot_result}. We find that the CLIP-based reward function completely fails to guide policy learning, while the RoboCLIP-based reward function demonstrates some capability for learning meaningful behaviors. Both RL-VLM-F and \ours{} successfully train policies, but \ours{} achieves the highest success rates, highlighting its superior ability to leverage VLM-generated preference signals for policy optimization in real-world scenarios.

\section{CONCLUSIONS}
We introduced \ours{}, a fully supervised, preference-based RL framework that bypasses the need for explicit reward modeling and human-annotated labels. By querying a large vision-language model (VLM) for pairwise preferences on visual trajectory segments, \ours{} trains policies using contrastive preference learning, effectively leveraging AI-generated feedback for agent training. Our experiments demonstrate that \ours{} matches or surpasses existing state-of-the-art VLM-based reward generation methods. While the real-world tasks in this study are relatively simple, they serve as a compelling proof of concept for the effectiveness of our approach. Future work can build on this foundation by extending \ours{} to more complex, long-horizon, and multi-stage tasks, further unlocking the potential of AI-generated feedback for scalable and robust robot learning.

% \newpage
\bibliographystyle{ieeetr} 
\bibliography{IEEEfull,references}

\addtolength{\textheight}{-12cm}   % This command serves to balance the column lengths
                                  % on the last page of the document manually. It shortens
                                  % the textheight of the last page by a suitable amount.
                                  % This command does not take effect until the next page
                                  % so it should come on the page before the last. Make
                                  % sure that you do not shorten the textheight too much.

%%%%%%%%%%%%%%%%%%%%%%%%%%%%%%%%%%%%%%%%%%%%%%%%%%%%%%%%%%%%%%%%%%%%%%%%%%%%%%%%

%%%%%%%%%%%%%%%%%%%%%%%%%%%%%%%%%%%%%%%%%%%%%%%%%%%%%%%%%%%%%%%%%%%%%%%%%%%%%%%%

%%%%%%%%%%%%%%%%%%%%%%%%%%%%%%%%%%%%%%%%%%%%%%%%%%%%%%%%%%%%%%%%%%%%%%%%%%%%%%%%
% \section*{APPENDIX}

% Appendixes should appear before the acknowledgment.

% \section*{ACKNOWLEDGMENT}

% The preferred spelling of the word ÒacknowledgmentÓ in America is without an ÒeÓ after the ÒgÓ. Avoid the stilted expression, ÒOne of us (R. B. G.) thanks . . .Ó  Instead, try ÒR. B. G. thanksÓ. Put sponsor acknowledgments in the unnumbered footnote on the first page.

\end{document}